\documentclass[11pt]{article}

\usepackage[final]{acl}

\usepackage{times}
\usepackage{latexsym}

\usepackage[T1]{fontenc}

\usepackage[utf8]{inputenc}

\usepackage{microtype}
\usepackage{algorithm}
\usepackage{algorithmic}
\usepackage{amsmath}
\usepackage{booktabs}
\usepackage{multirow}
\usepackage{multicol}
\usepackage{amssymb}   
\usepackage{pifont}
\newcommand{\xmark}{\ding{55}}
\usepackage{inconsolata}

\usepackage{graphicx}

\title{\textsc{CodMas}: A Dialectic Multi-Agent Collaborative Framework for Structured RTL Optimization}

\author{
 \textbf{Che-Ming Chang\thanks{Both authors contributed equally to this work. This work was conducted in part during an internship at IBM Research.}\textsuperscript{1}},
 \textbf{Prashanth Vijayaraghavan\footnotemark[1]\textsuperscript{2}},
 \textbf{Ashutosh Jadhav\textsuperscript{2}},
 \textbf{Charles Mackin\textsuperscript{2}},\\
 \textbf{Vandana Mukherjee\textsuperscript{2}},
 \textbf{Hsinyu Tsai\textsuperscript{2}},
 \textbf{Ehsan Degan\textsuperscript{2}}
\\\\
 \textsuperscript{1}National Taiwan University,
 \textsuperscript{2}IBM Research
\\
 \texttt{
   b09901156@ntu.edu.tw, prashanthv@ibm.com, ashutosh@us.ibm.com,}\\
   \texttt{charles.mackin@ibm.com, vandana@us.ibm.com,}\\
   \texttt{htsai@us.ibm.com, edehgha@us.ibm.com}
}

\begin{document}
\maketitle

\begin{abstract}

Optimizing Register Transfer Level (RTL) code is a critical step in Electronic Design Automation (EDA) for improving power, performance, and area (PPA). We present \textsc{CodMas} (\textit{Collaborative Optimization via a Dialectic Multi-Agent System}), a framework that combines structured dialectic reasoning with domain-aware code generation and deterministic evaluation to automate RTL optimization. At the core of \textsc{CodMas} are two dialectic agents: the \emph{Articulator}, inspired by rubber-duck debugging, which articulates stepwise transformation plans and exposes latent assumptions; and the \emph{Hypothesis Partner}, which predicts outcomes and reconciles deviations between expected and actual behavior to guide targeted refinements. These agents direct a Domain-Specific Coding Agent (DCA) to generate architecture-aware Verilog edits and a Code Evaluation Agent (CEA) to verify syntax, functionality, and PPA metrics. We introduce \textsc{RTLOpt}, a benchmark of 120 Verilog triples (unoptimized, optimized, testbench) for pipelining and clock-gating transformations. Across proprietary and open LLMs, \textsc{CodMas} achieves $\sim$25\% reduction in critical path delay for pipelining and $\sim$22\% power reduction for clock gating, while reducing functional and compilation failures compared to strong prompting and agentic baselines. These results demonstrate that structured multi-agent reasoning can significantly enhance automated RTL optimization and scale to more complex designs and broader optimization tasks.
\end{abstract}

\section{Introduction}

The increasing complexity of modern chip design has accelerated the integration of Artificial Intelligence (AI) into Electronic Design Automation (EDA) workflows, reducing reliance on manual effort and enabling faster design cycles. While AI has shown notable success in tasks such as logic synthesis~\cite{alphasyn} and placement~\cite{chipformer}, generating and optimizing Hardware Description Languages (HDLs) remains challenging. RTL (Register Transfer Level) code, typically written in Verilog or VHDL, requires careful hand-tuned transformations to meet power, performance, and area (PPA) constraints. Transformations such as pipelining and clock gating are particularly critical for performance and energy efficiency, yet they are time-consuming, error-prone, and demand deep domain expertise.
\begin{figure}[t]
    \centering
    \includegraphics[width=0.85\linewidth]{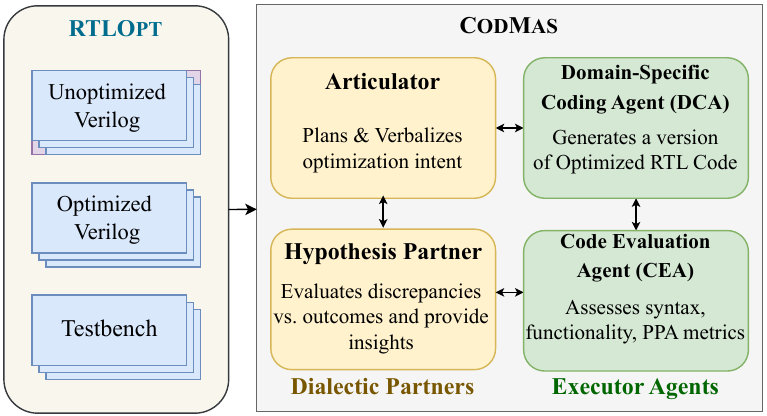}
    \caption{Overview of the \textsc{CodMas} framework, illustrating dialectic interaction between agents and iterative refinement of RTL code.}
    \label{fig:mascot}
\end{figure}
Commercial EDA tools provide automated support for certain RTL optimizations, but they lack explicit reasoning about design intent and early-stage architectural transformations. Existing learning-based approaches either focus on syntactic HDL generation~\cite{verigen, akyash2025rtl++, yu2025spec2rtl, RTLcoder, chipgpt, autochip, VerilogCoder} or rely on heuristic search or local rewriting~\cite{veriPPA, rtlrewriter, MCTS}, limiting their ability to generalize across designs and capture global architectural patterns. Moreover, datasets for RTL optimization remain scarce, limiting reproducibility, benchmarking, and systematic evaluation of learning-based methods.

To address these gaps, we introduce \textsc{RTLOpt}, a curated benchmark designed specifically for RTL-level optimization. It contains 120 Verilog code triples, each consisting of an unoptimized design, an optimized counterpart (via pipelining or clock gating), and an associated testbench. This organization enables both functional verification and quantitative evaluation of PPA metrics. The dataset spans diverse design patterns and complexity levels, providing a tractable yet representative foundation for research-scale experimentation and model evaluation. Building on this benchmark, we propose \textsc{CodMas}, a multi-agent framework for automated RTL optimization that integrates \emph{dialectic reasoning} into the optimization loop. The \emph{Articulator} agent verbalizes optimization intent, while the \emph{Hypothesis Partner} evaluates discrepancies between expected and actual outcomes. These reasoning agents coordinate with the \emph{Domain-Specific Coding Agent (DCA)} and the \emph{Code Evaluation Agent (CEA)} to iteratively refine RTL designs, maintaining functional correctness while improving PPA metrics. Figure~\ref{fig:mascot} illustrates this closed-loop interaction. Our key contributions include:

   \noindent \textbf{\textsc{RTLOpt}}: A benchmark optimization dataset of $\sim$120 Verilog-based pipelining and clock gating.
    
    \noindent \textbf{\textsc{CodMas}}: A multi-agent framework combining dialectic reasoning, domain-informed code generation, and PPA evaluation for RTL optimization.
    
    \noindent \textbf{Empirical validation}: Demonstrates consistent improvements across models and optimization scenarios, highlighting the efficacy of structured reasoning in automated RTL optimization.

\section{\textsc{RTLOpt}: HDL Optimization Dataset} \label{dataset}

To evaluate LLM performance in HDL code optimization, we collected data from GitHub repositories implementing optimization methods, focusing on Verilog. Since techniques like pipelining and clock gating are not always explicitly labeled, we used search terms like ``(pipelining OR clock gating) + verilog'' (e.g., ``pipelining verilog'' or ``pipelined verilog'') to identify relevant repositories. The data collection process was as follows: we filtered all publicly available repositories for self-contained Verilog files to reduce cross-file dependencies and simplify analysis. We then extracted accompanying testbenches and performed deduplication on the collected modules using token-level similarity and AST-structure hashing to reduce near-duplicates and mitigate overlap with public training corpora.

To specifically target pipelining and clock gating optimizations, we manually reviewed filtered Verilog files, inspecting code or descriptions indicating these optimization patterns. When optimization details were missing, we generated unoptimized versions (i.e., without pipelining or clock gating) and corresponding testbenches, either by modifying existing descriptions or editing LLM-generated code (e.g., DeepSeek models \cite{liu2024deepseek}). All generated or edited examples were subsequently validated through synthesis and simulation to ensure functional equivalence and measurable metric improvements, and were normalized to minimize stylistic cues between unoptimized and optimized pairs. Importantly, unoptimized versions were obtained independently rather than mechanically degraded from optimized code, avoiding reverse-engineering artifacts.

This process resulted in the \textsc{RTLOpt} dataset, consisting of triples of (unoptimized code, optimized code, testbench) for evaluation with designs spanning arithmetic, control, and mixed modules, ensuring diversity. While modest in scale (120 triples), \textsc{RTLOpt} is comparable to or larger than existing code evaluation benchmarks and is designed to prioritize realistic transformations and reproducible evaluation over raw size. The current form of the dataset focuses on pipelining and clock gating, but the dataset structure and tooling are designed to support future extensions to transformations such as retiming, resource sharing, and FSM restructuring. \textsc{RTLOpt} is a seed benchmark intended for community extension. See Table~\ref{tab:dataset} for dataset statistics.

\begin{table}[]


\centering
\small
\begin{tabular}{@{}lrl@{}}
\toprule
\multicolumn{3}{c}{\textbf{\textsc{RTLOpt} Dataset Statistics}}                        \\ \midrule
\# Pipelining Code Triples           & \multicolumn{2}{r}{70}                    \\
\# Clock Gating Code Triples         & \multicolumn{2}{r}{50}                    \\
Power Range $(nW)$             & \multicolumn{2}{r}{$\sim{1}-\sim{19,000}$}  \\

Area Range $(\mu m^2)$             & \multicolumn{2}{r}{$\sim{1}-\sim{18,000}$}  \\
Delay Range $(ns)$                & \multicolumn{2}{r}{$\sim{15}-\sim{1,600}$} \\ \bottomrule
\end{tabular}
\caption{Summary of the \textsc{RTLOpt} Dataset for evaluating pipelining and clock gating optimizations.}
\label{tab:dataset}
\end{table}

\begin{figure*}
    \centering
    \includegraphics[width=0.94\linewidth]{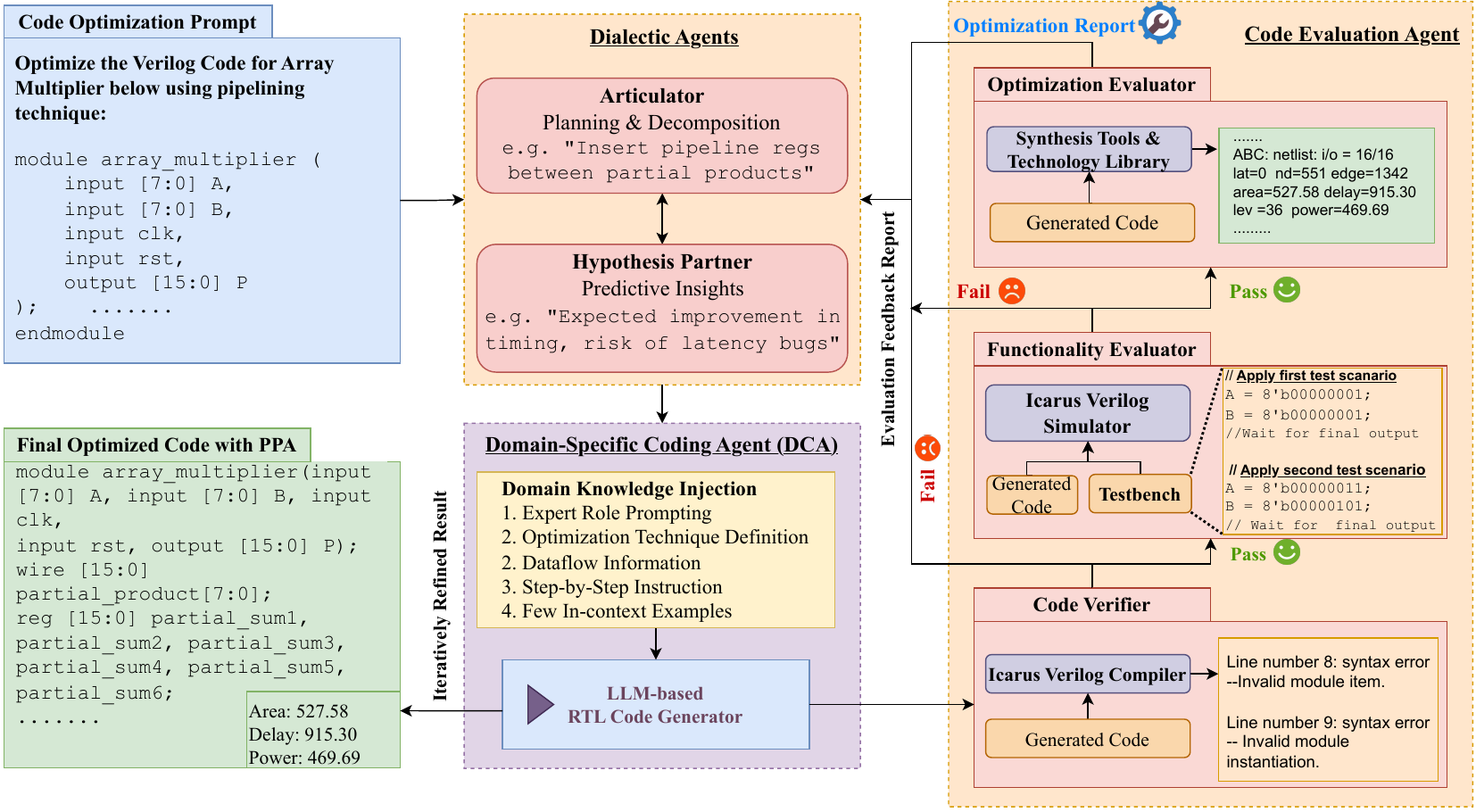}
    \caption{Illustration of the complete \textsc{CodMas} architecture, showing dialectic agents (Articulator \& Hypothesis Partner), the executor agents (Domain-Specific Coding Agent \& Code Evaluation Agent), and the iterative RTL refinement loop (e.g., pipelined array multiplier). The example highlights a pipelined array multiplier optimization, with feedback from different components guiding successive refinements toward improved PPA metrics.} 
    \label{fig:mascot_arch}
\end{figure*}

\section{Methodology}

We introduce \textsc{CodMas} (\textit{Collaborative Optimization via Dialectic Multi-Agent System for Structured RTL}), a multi-agent framework for automated RTL optimization. \textsc{CodMas} is grounded on two insights: (1) iterative, structured reasoning between complementary agents enhances optimization quality, and (2) domain-specific architectural knowledge must guide LLM-based RTL transformations. The system comprises four specialized agents: two \emph{dialectic reasoning agents} (Articulator and Hypothesis Partner) and two \emph{executor agents} (Domain-Specific Coding Agent (DCA), and Code Evaluation Agent (CEA)). Unlike conventional LLM-based pipelines, \textsc{CodMas} integrates planning, hypothesis formation, code generation, and deterministic evaluation in a closed feedback loop. This structure supports interpretable, performance-aware refinement and can generalize beyond the initial 120 Verilog triples in \textsc{RTLOpt}, allowing application to larger designs, cross-file modules, and diverse RTL transformations.

\subsection{Dialectic Reasoning Agents}

The dialectic reasoning agents in \textbf{\textsc{CodMas}} operate as complementary collaborators within a structured, pair programming-inspired optimization workflow. Unlike typical single-agent or monolithic reasoning pipelines, RTL optimization benefits from explicitly separating \emph{design articulation} from \emph{hypothesis generation}. Accordingly, both agents engage in reflective design thinking but assume distinct cognitive roles: the \textbf{Articulator} focuses on decomposition and planning, while the \textbf{Hypothesis Partner} specializes in predictive reasoning and diagnostic inference. Their iterative interaction forms a closed-loop dialectic, where articulation, critique, and hypothesis refinement guide the executor agents toward functionally correct and performance-optimized RTL.

\subsubsection{Articulator}

The Articulator serves as the planning and verbalization module of the system. Inspired by \textit{rubber-duck debugging} and structured reasoning, it analyzes the unoptimized RTL design then generates a stepwise transformation plan aligned with target optimization objectives, such as pipelining or clock gating. Each step is interpretable and designed to preserve functional correctness, enabling traceable optimization workflows. Procedural reasoning allows the Articulator to decompose complex transformations into ordered operations while surfacing latent design assumptions. These assumptions are expressed in a structured form consumable by both human designers and executor agents. For instance, it may recommend inserting pipeline registers between critical computation stages or adjusting delay alignment logic to preserve correctness. This articulated plan provides a shared semantic scaffold for hypothesis formation, code generation, and iterative refinement, remaining largely stable across hypothesis proposals to enable consistent evaluation of alternative transformations under a unified circuit model.

\subsubsection{Hypothesis Partner}

The Hypothesis Partner operates in tandem with the Articulator, leveraging \textit{abductive} and \textit{model-based reasoning} to anticipate the effects of planned transformations on functional correctness and PPA metrics. Unlike conventional code-generation approaches, it begins reasoning before any optimized code is produced, formulating hypotheses conditioned on the Articulator's performance-annotated circuit representation. Given a fixed structural interpretation, the Hypothesis Partner proposes metric-targeted transformations, predicts performance gains, identifies functional or structural pitfalls, and guides corrective strategies. When synthesis or simulation feedback reveals deviations, the agent revises prior hypotheses, attributes failure causes, and suggests targeted refinements. The iterative reasoning loop embeds verification-awareness into the optimization process, ensuring that code generation is guided by actionable, performance-aligned insights. This separation enables systematic exploration of multiple candidate transformations under a shared design model, avoiding premature commitment to suboptimal directions. As shown in Section~\ref{sec:ablations}, collapsing these roles into a unified agent degrades convergence stability and final metric gains, supporting the necessity of this dialectic structure for RTL optimization.

\subsection{Executor Agents}

Executor agents in \textbf{\textsc{CodMas}} operationalize the dialectic agents' reasoning by generating candidate RTL and providing rigorous feedback. While dialectic agents handle planning and predictive inference, executor agents convert these insights into actionable code and deterministic evaluation. This layer comprises the \textit{Domain-Specific Coding Agent (DCA)} for optimized RTL generation and the \textit{Code Evaluation Agent (CEA)} for syntax, functional, and PPA assessment. Together, they form a tightly coupled loop that iteratively refines RTL designs.

\subsubsection{Domain-Specific Coding Agent (DCA)}

The DCA translates the Articulator’s transformation plan and the Hypothesis Partner’s predicted outcomes into first-pass optimized Verilog code. It employs a multi-faceted prompt strategy via the \textit{Domain Knowledge Injector (DKI)}: (a) it positions the LLM as an expert RTL designer instructed to apply optimization techniques such as pipelining or clock gating (\textit{role prompting}); (b) embeds $K$ annotated examples of unoptimized and optimized Verilog code aligned with the articulated plan and predicted outcomes (\textit{example-based guidance}); and (c) incorporates structural context from a dataflow graph extracted with Pyverilog (\textit{tool-informed context}). Rather than providing raw RTL alone, the injected dataflow graph encodes register stages, combinational logic cones, control dependencies, and clock enables in a constrained schema, enabling the model to reason explicitly about pipeline depth, stage imbalance, and safe gating opportunities. For example, in a pipelining task, the graph exposes a long combinational path between two registers, guiding the insertion of intermediate registers while preserving control alignment.

Using this enriched prompt, the DCA generates a functionally equivalent, architecture-aware candidate design. The CEA evaluates this code, and feedback is returned to the dialectic agents: the Articulator revises transformation steps, the Hypothesis Partner updates predicted outcomes, and the DCA incorporates these adjustments into the next iteration. This cycle repeats until functional correctness and target PPA improvements are achieved.



\subsubsection{Code Evaluation Agent (CEA)}

The CEA provides deterministic, verifiable evaluation using open-source EDA tools (Icarus Verilog for compilation, GTKWave for simulation, and Yosys for synthesis and timing analysis). It evaluates designs along three dimensions: syntactic correctness, functional equivalence, and optimization quality. The \textit{Code Verifier} ensures compilable Verilog and produces structured error messages to guide correction; the \textit{Functionality Evaluator} simulates designs with testbenches adjusted for transformations such as pipelining to detect behavioral mismatches or regressions; and the \textit{Optimization Evaluator} synthesizes designs to extract PPA metrics for performance-aware refinement. The CEA compiles a unified feedback report consumed by the dialectic agents: syntactic errors prompt revision of transformation plans, functional mismatches trigger updates to assumptions, and suboptimal PPA metrics guide optimization strategy adjustments. The updated plans and hypotheses are sent back to the DCA, which generates refined RTL code. 

\subsection{Integrated Optimization Workflow}

Algorithm~\ref{alg:mascotdp} summarizes the \textsc{CodMas} optimization pipeline. Given unoptimized RTL $C_0$, target goals $G$ (e.g., PPA improvements), and iteration limit $T$, the \textit{Articulator} generates a structured transformation plan while the \textit{Hypothesis Partner} predicts expected functional behavior and PPA outcomes. A dataflow DAG supports structural reasoning, and the \textit{Domain Knowledge Injector} composes a prompt combining plan, hypotheses, and structural context for the LLM to generate candidate optimized RTL. The \textit{Code Evaluation Agent (CEA)}, comprising the Code Verifier, Functionality Evaluator, and Optimization Evaluator, assesses syntax, functional equivalence, and PPA. Feedback is incorporated into the dialectic loop: syntactic errors adjust the plan, functional mismatches update assumptions, and suboptimal PPA metrics guide strategy refinements. This loop continues until optimization goals are met or the iteration limit is reached, providing a traceable, verification-aware, and performance-driven optimization workflow.

\begin{table*}[]
\small
\centering
\begin{tabular}{@{}l|ccc|ccc|ccc|ccc@{}}
\toprule
\multicolumn{13}{c}{\textbf{Pipelining}} \\ \midrule
\textbf{Models} & \multicolumn{3}{c|}{\textbf{Zero-Shot}} & \multicolumn{3}{c|}{\textbf{Prompting/Agentic}} & \multicolumn{3}{c|}{\textbf{LLM-VeriPPA}} & \multicolumn{3}{c}{\textbf{\textsc{CodMas}}} \\ \cmidrule(lr){2-4} \cmidrule(lr){5-7} \cmidrule(lr){8-10} \cmidrule(lr){11-13}
 & A ($\downarrow$) & T ($\uparrow$) & FR ($\downarrow$) & A ($\downarrow$) & T ($\uparrow$) & FR ($\downarrow$) & A ($\downarrow$) & T ($\uparrow$) & FR ($\downarrow$) & A ($\downarrow$) & T ($\uparrow$) & FR ($\downarrow$) \\ \midrule
GPT-4o         & \textbf{0.985} & \textbf{11.4} & \textbf{54.2} & \textbf{0.982} & \textbf{15.2} & \textbf{49.6} & \textbf{0.986} & \textbf{15.4} & \textbf{39.7} & \textbf{0.960} & \textbf{25.5} & \textbf{19.5} \\
GPT-3.5-turbo  & 0.989 & 10.0 & 58.0 & 0.988 & 14.0 & 51.2 & 1.008 & 13.6 & 43.3 & 0.972 & 21.3 & 23.4 \\
DeepSeek-v2.5  & 1.009 & 9.8  & 57.8 & 0.986 & 12.8 & 50.0 & 1.025 & 13.0 & 42.8 & 0.979 & 21.4 & 22.8 \\
Llama-3        & 1.045 & 8.7  & 61.4 & 1.181 & 11.6 & 53.8 & 1.116 & 11.3 & 47.1 & 1.032 & 19.8 & 25.5 \\\midrule
Granite-34b    & \textbf{1.026} & 3.9  & 65.2 & \textbf{1.015 }& 5.3  & \textbf{59.1} & \textbf{1.022} & 6.6  & 57.7 & \textbf{0.998} & 10.5 & \textbf{28.3} \\
CodeLlama-34b  & 1.035 & \textbf{4.2}  & \textbf{64.7} & 1.036 & \textbf{6.8 } & 60.7 & 1.039 & \textbf{7.2}  & \textbf{56.1} & 1.030 & \textbf{11.2} & 29.5 \\ \midrule\midrule
\multicolumn{1}{l|}{Human} & \multicolumn{3}{c|}{N/A} & \multicolumn{3}{c|}{N/A} & \multicolumn{3}{c|}{N/A} & \textbf{0.848} & \textbf{45.6} & \textbf{0.0} \\ \midrule

\multicolumn{13}{c}{\textbf{Clock Gating}} \\ \midrule
\textbf{Models} & \multicolumn{3}{c|}{\textbf{Zero-Shot}} & \multicolumn{3}{c|}{\textbf{Prompting/Agentic}} & \multicolumn{3}{c|}{\textbf{LLM-VeriPPA}} & \multicolumn{3}{c}{\textbf{\textsc{CodMas}}} \\ \cmidrule(lr){2-4} \cmidrule(lr){5-7} \cmidrule(lr){8-10} \cmidrule(lr){11-13}
 & A ($\downarrow$) & P ($\uparrow$) & FR ($\downarrow$) & A ($\downarrow$) & P ($\uparrow$) & FR ($\downarrow$) & A ($\downarrow$) & P ($\uparrow$) & FR ($\downarrow$) & A ($\downarrow$) & P ($\uparrow$) & FR ($\downarrow$) \\ \midrule
GPT-4o         & \textbf{1.023} & \textbf{7.8} & \textbf{52.5} & \textbf{1.010} & \textbf{9.1} & \textbf{46.8} & \textbf{1.009} & \textbf{9.3} & \textbf{38.9} & \textbf{0.999} & \textbf{21.7} & \textbf{21.8} \\
GPT-3.5-turbo  & 1.035 & 6.3  & 55.6 & 1.018 & 7.5  & 49.6 & 1.019 & 7.9 & 43.4 & 1.020 & 18.8 & 24.2 \\
DeepSeek-v2.5  & 1.030 & 6.2  & 54.8 & 1.014 & 7.8  & 48.9 & 1.020 & 8.2 & 42.9 & 1.015 & 19.0 & 23.6 \\
Llama-3        & 1.093 & 5.5  & 58.6 & 1.060 & 6.7  & 52.1 & 1.062 & 7.0 & 46.5 & 1.048 & 16.5 & 26.3 \\\midrule
Granite-34b    & \textbf{1.037} & 3.0  & 63.0 & \textbf{1.048} &\textbf{ 5.8}  & 59.8 & \textbf{1.049 }& \textbf{5.9}  & \textbf{55.8} & \textbf{1.030} & \textbf{12.9} & \textbf{29.5} \\
CodeLlama-34b  & 1.054 & \textbf{3.4 } & \textbf{61.6} & 1.053 & 5.5  & \textbf{58.7} & 1.050 & 5.4  & 55.3 & 1.042 & 10.6 & 31.3 \\ \midrule\midrule
\multicolumn{1}{l|}{Human} & \multicolumn{3}{c|}{N/A} & \multicolumn{3}{c|}{N/A} & \multicolumn{3}{c|}{N/A} & \textbf{0.925} & \textbf{30.4} & \textbf{0.0} \\  
\bottomrule
\end{tabular}
\caption{Performance comparison on pipelining and clock gating (CG). Pipelining reports {Area (A)}, {Timing (T)}, and {Failure Rate (FR)}; clock gating reports {A}, {Power (P)}, and {FR}. Bold indicates best; $\uparrow$ higher is better, $\downarrow$ lower is better. All improvements are statistically significant ($p<0.01$) via paired t-tests with  standard deviation in Table \ref{tab:codmas_mean_std}.}


\label{tab:results}
\end{table*}






\section{Experiments}
\label{experiment}

We evaluate \textsc{CodMas} on RTL optimization, presenting the experimental setup, baselines, and metrics. Our study addresses three questions: \textbf{RQ1: Effectiveness of \textsc{CodMas}:} How does the framework improve PPA compared to other baselines? \textbf{RQ2: Component-wise Contribution:} What is the impact of the dialectic and executor agents on optimization performance? \textbf{RQ3: Benefits of Iterative Refinement:} How do multi-step reasoning and iterative refinement affect convergence speed and optimization quality across designs?

\paragraph{Baselines}
\label{sec:baselines}

We compare \textsc{CodMas} against representative prompting and agent-based approaches for RTL optimization. Our baselines include \textbf{Zero-Shot} prompting, intermediate-reasoning methods such as \textbf{CoDes}~\cite{vijayaraghavan2024chain}, \textbf{ReAct}~\cite{yao2023react}, and \textbf{Reflexion}~\cite{Reflexion2023}, and the two-stage correction–optimization pipeline \textbf{LLM-VeriPPA}~\cite{veriPPA}. All systems are evaluated using identical simulation and synthesis flows across proprietary models (GPT-4o, GPT-3.5-turbo) and open-source models (Llama-3, DeepSeek-v2.5, Granite-34B-Code, CodeLLaMA-34B). Detailed baseline configurations appear in the Appendix~\ref{appendix:baselines}. Although baselines such as ReAct, Reflexion, and CoDes were originally developed for general software reasoning tasks, we adapt them for RTL optimization by providing the same RTL parsing, transformation, and synthesis-feedback interfaces as used in \textsc{CodMas}. This ensures a fair comparison while preserving the original reasoning strategies of these models within the hardware optimization context.

\paragraph{Metrics}  
\label{sec:metrics}

We evaluate optimized RTL relative to the original design using synthesis and simulation with Yosys and Liberty-based standard-cell libraries. Area (\(A\)) is reported as \(A/A_0\), where \(A_0\) is the baseline, with values below 1 indicating reduction. Power (\(P\)) and timing (\(T\)) improvements are expressed as percentage gains, with timing measured via critical path delay (CPD), where positive gains indicate reduced CPD. Failure rate (FR) captures the fraction of designs that fail functional, synthesis, or timing checks after optimization, with lower FR indicating higher reliability (see Appendix \ref{app:metrics}). All optimized designs are validated using module-level testbenches derived from original repositories or constructed from behavioral specifications. While we do not yet incorporate formal equivalence checking, our methodology aligns with common industrial RTL optimization pipelines, where simulation-based testing remains one of the key validation mechanisms.


\section{Results}
\subsection{(RQ1) Effectiveness of \textsc{CodMas} }
Table~\ref{tab:results} summarizes PPA outcomes for six LLMs under four optimization strategies. Across all models, \textsc{CodMas} consistently delivers the strongest improvements: pipelining achieves timing gains above $20\%$ (reaching $25.5\%$ on GPT-4o), while clock gating attains power reductions exceeding $19\%$ on average, compared to less than $10\%$ for all baselines. Failure rates under \textsc{CodMas} remain below $30\%$, while prompting and agentic baselines typically exceed $40\%$ to $50\%$, indicating more frequent syntax, functional, and PPA violations. Model-wise trends reinforce these findings. GPT-4o leads across strategies, achieving $\sim25\%$ timing improvement in pipelining and $\sim22\%$ power reduction in clock gating with FR below $25\%$. Open-source models Granite-34b and CodeLlama-34b perform poorly under zero-shot or naive prompting, with minimal PPA gains and FR above $60\%$, yet under \textsc{CodMas} they reach up to $13\%$ timing and power improvements with FR under $30\%$. Llama3 is the most challenging: baseline modes increase area up to $18\%$ with FR above $50\%$, but \textsc{CodMas} reduces net area impact to $A/A_0 \approx 1.03$ while achieving timing gains near $20\%$. Error analysis indicates baseline failures arise from syntax ($\sim40\%$), functional ($\sim35\%$), and PPA ($\sim25\%$) issues, all mitigated by \textsc{CodMas} through explicit planning, hypothesis-guided reasoning, and deterministic evaluation. Overall, \textsc{CodMas} consistently improves PPA, lowers FR, and stabilizes optimization across both proprietary and open-source LLMs.

\subsubsection{Impact on Area}
Pipelining adds registers and handshake logic, and clock gating adds gating cells and control logic, often increasing area despite timing or power gains. In our experiments, \textsc{CodMas} keeps area near baseline, with GPT-4o showing $\sim4\%$ reductions under both optimizations while achieving notable timing or power improvements. Area changes are generally modest, due to synthesis variations or minor restructuring. Baselines typically show minimal area change with limited PPA gains; for example, Llama-3 in zero-shot pipelining increases area $\sim4.5\%$ with only $\sim8.7\%$ timing gain and high FR ($61.4\%$). The Articulator's transformations and Hypothesis Partner's forecasts, validated by the CEA, focus on PPA improvements while keeping area changes secondary.

\subsection{(RQ2) Component-wise Contribution}
\label{sec:ablations}

Table~\ref{tab:ablations} presents an ablation study of three key \textsc{CodMas} components: Dialectic Agents (DA), Domain Knowledge Injection (DKI), and the Code Evaluation Agent (CEA), reporting metrics for GPT-4o and Llama-3. The complete \textsc{CodMas} pipeline consistently outperforms all ablated variants, demonstrating the importance of each module: DA coordinates structured reasoning, DKI grounds transformations in design intent, and CEA filters invalid or low-quality edits.  

Removing the Dialectic Agents (\textit{w/o DA}) results in the largest drop in performance. Timing gains for GPT-4o pipelining fall from $25.5\%$ to $12.9\%$, and failure rates rise to $38.5\%$--$44.7\%$, highlighting DA’s role in structured refinement and hypothesis-guided filtering. Without CEA (\textit{w/o CEA}), failure rates increase (e.g., $21.8\%$ to $32.7\%$ for GPT-4o clock gating), as flawed edits persist. Omitting DKI (\textit{w/o DKI}) reduces optimization quality and robustness, particularly for smaller models: Llama-3 pipelining timing gains drop from $19.8\%$ to $9.3\%$, and FR rises by $\sim10$ points. These results confirm that each component (DA, DKI, and CEA) provides complementary benefits that are crucial for achieving stable RTL optimization.

\begin{table}[]
\small
\centering
\begin{tabular}{@{}lcccc@{}}
\toprule
\textbf{} & \textbf{\textsc{Full}} & \textbf{w/o CEA} & \textbf{w/o DA} & \textbf{w/o DKI} \\ \midrule
\multicolumn{5}{c}{\textbf{Pipelining}} \\ \midrule
T (GPT-4o)            & \textbf{25.5} & 17.2 & 12.9 & 11.5 \\
FR (GPT-4o)           & \textbf{19.5} & 31.0 & 38.5 & 32.3 \\
T (Llama-3)          & \textbf{19.8} & 12.6 & 10.4 & 9.3 \\
FR (Llama-3)         & \textbf{25.5} & 36.9 & 44.7 & 35.8 \\\midrule
\multicolumn{5}{c}{\textbf{Clock Gating}} \\ \midrule
P (GPT-4o)            & \textbf{21.7} & 14.4 & 9.5 & 11.1 \\
FR (GPT-4o)           & \textbf{21.8} & 32.7 & 37.2 & 40.5 \\
P (Llama-3)          & \textbf{16.5} & 10.2 & 6.8 & 7.2 \\
FR (Llama-3)         & \textbf{26.3} & 35.5 & 41.3 & 43.2 \\ \bottomrule
\end{tabular}
\caption{Component-wise ablation study of \textsc{CodMas}. DA: Dialectic Agents; DKI: Domain Knowledge Injection; CEA: Code Evaluation Agent. }
\label{tab:ablations}
\end{table}

\subsubsection{Dialectic Agent Ablation}
\label{sec:dialectic_ablation}

To isolate the impact of the dialectic agent design in \textsc{CodMas}, we compare the full system against a single alternative architecture: a shared-memory multi-agent (SMA) variant in which the Articulator and Hypothesis Partner roles are collapsed, and both agents jointly interpret and modify the RTL without explicit separation of planning and predictive reasoning. All variants are matched for synthesis calls to ensure fair comparison.





\begin{figure}
    \centering
    \includegraphics[width=\linewidth]{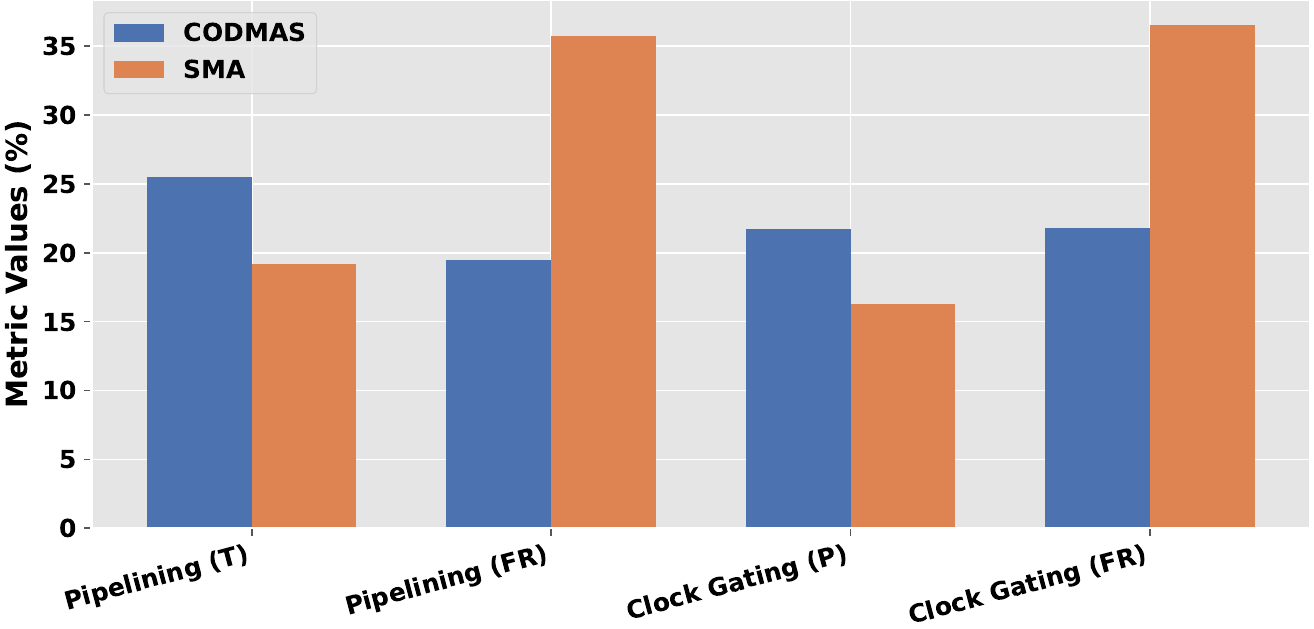}
    \caption{Ablation comparing CODMAS with a shared-memory multi-agent (SMA) variant where the Articulator and Hypothesis Partner roles are combined.}
    \label{fig:dialectic_ablations}

    \label{fig:easy_vs_hard}
\end{figure}

Results in Figure~\ref{fig:dialectic_ablations} show that collapsing the two roles into a shared-memory multi-agent system consistently reduces performance gains and increases failure rates. For instance, pipelining timing improvements drop from $25.5\%$ to $19.2\%$ for GPT-4o, and failure rates increase from $19.5\%$ to $35.7\%$. Similarly, clock gating power gains decrease and FR rises. These findings demonstrate that the explicit separation between the Articulator and Hypothesis Partner is critical: articulated planning provides a stable semantic scaffold, while the predictive hypotheses guide targeted transformations. Together, this structure enables higher metric gains, and more reproducible RTL optimization.

\subsection{(RQ3) Benefits of Iterative Refinement}
\begin{figure}
    \centering
    \includegraphics[width=0.99\linewidth]{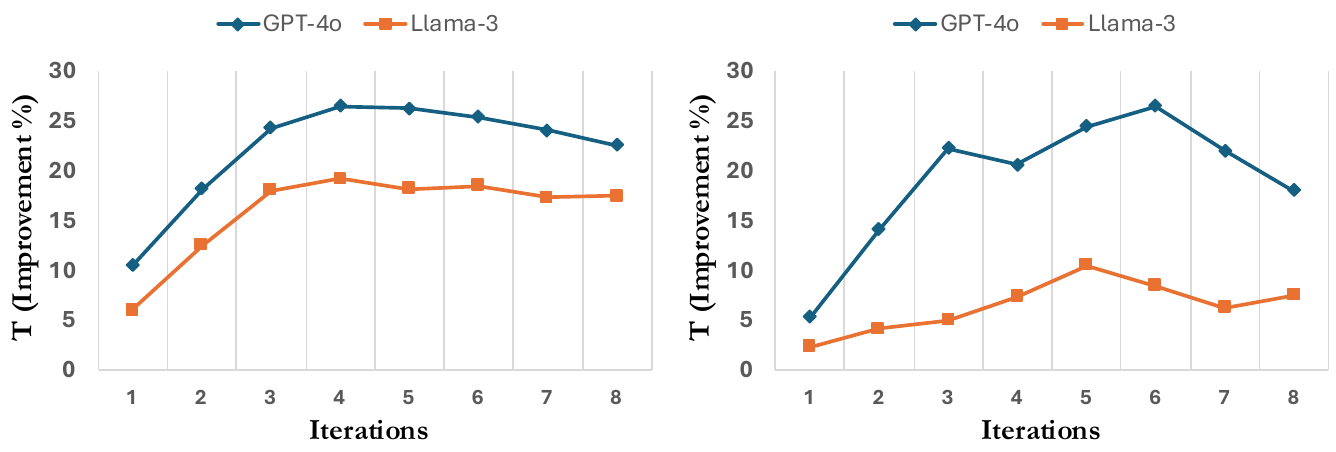}
    \caption{Impact of iterative refinement in \textsc{CodMas} on pipelining timing improvement (\%) for easy (left) and hard (right) \textsc{RTLOpt} problems. Early iterations yield substantial gains; later iterations show diminishing or unstable returns, especially for harder cases.}

    \label{fig:easy_vs_hard}
\end{figure}
\label{sec:iterative}
To assess the impact of iterative feedback on pipelining, we evaluate \emph{easy} and \emph{hard} tasks over eight refinement iterations (Figure~\ref{fig:easy_vs_hard}). Both scenarios show steep gains in the first three iterations, indicating that initial feedback captures the largest optimization opportunities. Easy tasks plateau by iteration 4 and decline slightly thereafter, while hard tasks peak around iterations 5-6 and then oscillate, reflecting dense pipelines and conflicting transformation hypotheses. These trends highlight the value of iterative refinement for systematically improving performance, while also indicating that excessive iterations may yield marginal or unstable gains. Adaptive stopping or dynamic iteration strategies can mitigate wasted computation and prevent regressions in complex designs.



\section{Conclusion}
\label{sec:conclusion}
Achieving efficient power, performance, and area (PPA) in RTL designs is a critical yet challenging task in modern hardware design. To address this, we introduce \textsc{RTLOpt}, a benchmark for pipelining and clock-gating optimizations, and \textsc{CodMas}, a multi-agent framework combining dialectic reasoning with domain-specific code generation and deterministic evaluation. The Articulator and Hypothesis Partner guide executor agents (Domain-Specific Coding Agent and Code Evaluation Agent) to produce and assess Verilog designs rigorously. Our experiments demonstrate $\sim25\%$ timing improvement and $\sim22\%$ power reduction with failure rates $<30\%$, with ablations showing all components and iterative refinement are essential for robust performance. Future directions include expanding the dataset, exploring adaptive iteration strategies, extending to additional RTL optimizations, leveraging retrieval-augmented prompting, full synthesis flow validation, and incorporating self-play or reinforcement learning to further enhance optimization outcomes.


\section*{Limitations}

While our framework advances automated RTL optimization, several limitations remain. First, \textsc{CodMas} has been evaluated primarily on pipelining and clock-gating transformations, and its generalization to broader categories of RTL optimizations (e.g., retiming, resource sharing, FSM restructuring) is not yet fully established. Second, although the evaluation pipeline incorporates deterministic EDA tools, scalability to very large industrial designs is constrained by tool runtime and the need for repeated synthesis queries. Third, the dialectic reasoning agents occasionally generate overly generic transformation plans that require iterative refinement, indicating that the system still relies on principled prompting and task-specific templates. Fourth, \textsc{RTLOpt} is a seed benchmark with limited size, and expanding it to capture the diversity of industrial RTL coding styles and multi-file hierarchies is an important direction for future work. Finally, because functional equivalence is verified using standard testbenches rather than exhaustive formal techniques, subtle corner-case divergences may go undetected in rare scenarios. These limitations highlight opportunities for improving reasoning robustness, dataset coverage, and scalability in future iterations of the framework.

\section*{Ethical Considerations}
Although \textsc{RTLOpt} is built entirely from publicly licensed Verilog code, integrating it into an automated RTL‑optimization workflow introduces certain security considerations. Prior work has demonstrated that LLM‑generated RTL can contain vulnerabilities cataloged under Common Weakness Enumerations (CWEs)~\cite{gadde2024all}. Furthermore, LLMs for HDL generation may be susceptible to data‑poisoning or backdoor attacks, where compromised training data leads to the generation of insecure or malicious circuit components~\cite{mankali2025rtl}. To mitigate these risks, our system emphasizes human oversight and interpretability by generating explicit transformation plans and hypotheses that expert designers can review and approve. We also perform rigorous simulation and synthesis checks to ensure deterministic validation and detect unintended structural or security flaws. All modules in \textsc{RTLOpt} are fully documented with origin and licensing information to support clear provenance tracking. Finally, we recommend that any deployment of automatically optimized hardware include additional security audits, formal verification, and human review, particularly in safety or security-critical applications.

\bibliography{main}
\newcommand{\cmark}{\ding{51}}
\appendix
\newpage

\section{Algorithm: \textsc{CodMas} }

\begin{algorithm}[H]
\caption{\textsc{CodMas} Optimization Loop}
\label{alg:mascotdp}
\begin{algorithmic}[1]
\REQUIRE Input RTL $C_0$, Opt. Goal $G$, Iteration Cap $T$
\STATE $P \leftarrow \texttt{ArticulatorInit}(C_0, G)$
\STATE $H \leftarrow \texttt{HypoPartnerInit}(C_0, G)$
\STATE $D \leftarrow \texttt{DataflowGraph}(C_0)$
\STATE $\Pi \leftarrow \texttt{DKI}(P, H, D)$
\STATE $C \leftarrow \texttt{LLMGenerate}(C_0, \Pi)$
\STATE $(E_{\text{syn}}, E_{\text{func}}, M_{\text{ppa}}) \leftarrow \texttt{CEA}(C, C_0)$
\STATE $t \leftarrow 0$
\WHILE{$(E_{\text{syn}} \ne \emptyset$ \OR $E_{\text{func}} \ne \emptyset$ \OR $M_{\text{ppa}} \not\ge G)$ \AND $t < T$}
    \IF{$E_{\text{syn}} \ne \emptyset$}
        \STATE $E_{\text{func}} \leftarrow \emptyset$, $M_{\text{ppa}} \leftarrow \emptyset$
        \STATE $P \leftarrow \texttt{ArticulatorUpdate}(P, E_{\text{syn}})$
    \ELSIF{$E_{\text{func}} \ne \emptyset$}
        \STATE $M_{\text{ppa}} \leftarrow \emptyset$
        \STATE $H \leftarrow \texttt{HypoPartnerUpdate}(H, E_{\text{func}})$
        \STATE $P \leftarrow \texttt{ArticulatorAssist}(P, E_{\text{func}})$
    \ELSIF{$M_{\text{ppa}} \not\ge G$}
        \STATE $P \leftarrow \texttt{ArticulatorUpdate}(P, M_{\text{ppa}})$
        \STATE $H \leftarrow \texttt{HypoPartnerUpdate}(H, M_{\text{ppa}})$
    \ENDIF
    \STATE $\Pi \leftarrow \texttt{DKI}(P, H, D)$
    \STATE $C \leftarrow \texttt{LLMGenerate}(C_0, \Pi)$
    \STATE $(E_{\text{syn}}, E_{\text{func}}, M_{\text{ppa}}) \leftarrow \texttt{CEA}(C, C_0)$
    \STATE $t \leftarrow t + 1$
\ENDWHILE
\RETURN Final optimized RTL $C$
\end{algorithmic}
\end{algorithm}

\section{Metric Computation Details}
\label{app:metrics}

Optimized RTL designs are evaluated along four key dimensions: Area (A), Power (P), Timing (T), and Failure Rate (FR). All metrics are computed using Yosys with Liberty-based standard-cell libraries.

\paragraph{Area (A):} Computed as the total synthesized cell area. Reported relative to baseline as $A / A_0$, where $A_0$ is the unoptimized design. Values $<1$ indicate area reduction, while $>1$ indicate an increase.

\paragraph{Power (P):} Estimated static and dynamic power consumption using Yosys synthesis reports. Percentage improvement is computed as
\[
P_{\%} = \frac{P_0 - P}{P_0} \times 100
\]
where $P_0$ is the power of the original RTL.

\paragraph{Timing (T):} Measured by critical path delay (CPD), defined as the longest combinational path delay through library cells. Percentage improvement is
\[
T_{\%} = \frac{CPD_0 - CPD}{CPD_0} \times 100
\]
where $CPD_0$ is the baseline.

\paragraph{Failure Rate (FR):} Fraction of RTL designs that fail verification or synthesis checks after optimization. A design is considered failed if it violates functional correctness, fails synthesis, or exceeds timing constraints. Lower FR indicates higher reliability and robustness of the optimization framework. All metrics are reported per design, with averages and standard deviations provided across the dataset for aggregate evaluation. CPD is adjusted for structural transformations such as pipelining (accounting for latency shifts) to ensure fair comparison. Power and area are normalized by baseline RTL to facilitate cross-design comparison.

\section{Datasets}

To evaluate automated RTL optimization, we introduce \textsc{RTLOpt}, a dataset designed for reproducible and metric-driven assessment. Table~\ref{tab:verilog_datasets} compares \textsc{RTLOpt} against prior Verilog datasets. Existing benchmarks such as VerilogEval~\cite{verilogeval} and TuRTLe~\cite{garcia2025turtle} focus on functional correctness but lack synthesizability or metric-specific evaluation, limiting their utility for evaluating PPA-aware transformations. Other datasets, including RTLRewriter~\cite{rtlrewriter} and ResBench~\cite{guo2025resbench}, provide synthesizable RTL but do not include functional tests or metric-oriented targets, restricting systematic assessment of optimization performance.

\textsc{RTLOpt} contains 120 Verilog triples (unoptimized, optimized, and testbench), all synthesizable and functionally validated, with clearly defined PPA objectives such as pipelining and clock gating. This allows rigorous evaluation of both correctness and optimization effectiveness, enabling quantitative comparisons across LLM-driven frameworks. By explicitly including metric-specific targets, \textsc{RTLOpt} fills a critical gap, providing a standardized benchmark for evaluating end-to-end RTL optimization pipelines in a reproducible manner.

\begin{table*}[ht]
\centering
\small
\begin{tabular}{lcccc}
\toprule
\textbf{Dataset} & \textbf{Size} & \textbf{Functionality} & \textbf{Synthesizability} & \textbf{Metric-specific} \\
\midrule
VerilogEval~\cite{verilogeval} & 156 & \cmark & \xmark & \xmark \\
RTLLM~\cite{rtllm} & 30 & \cmark & \cmark & \xmark  \\
RTLRewriter~\cite{rtlrewriter} & 95 & \xmark & \cmark & \xmark  \\
TuRTLe~\cite{garcia2025turtle} & 223 & \cmark & \cmark & \xmark  \\
CVDP~\cite{pinckney2025comprehensive} & 783 & \cmark & \cmark & \xmark  \\
ResBench~\cite{guo2025resbench} & 56 & \cmark & \cmark & \xmark \\ \midrule
\textbf{\textsc{RTLOpt} (Ours)} & 120 & \cmark & \cmark & \cmark  \\
\bottomrule
\end{tabular}
\caption{Comparison of \textsc{RTLOpt} with prior Verilog datasets. ``Metric-specific'' indicates whether metric-specific optimization exists or not for evaluation.}
\label{tab:verilog_datasets}
\end{table*}

\begin{table*}[h!]
\centering
\small
\begin{tabular}{@{}lcc@{}}
\toprule
\textbf{Module} & \textbf{Original RTL} & \textbf{\textsc{CodMas} Optimized RTL} \\ \midrule
Multiplier      & Sequential multiply w/o pipeline & Partial product pipelined with inter-stage registers \\
Adder           & Ripple-carry 32-bit adder & Pipelined adder with reduced critical path \\
Control Logic   & Unconditioned enable signals & Handshake-aware control signals with proper gating \\ \midrule
\textbf{Observed Errors in Baselines} & \multicolumn{2}{c}{Syntax errors, functional mismatches, unmet PPA targets} \\
\textbf{Corrected by \textsc{CodMas}} & \multicolumn{2}{c}{All syntax errors fixed, functional simulation passes, critical path reduced $\sim25\%$} \\
\bottomrule
\end{tabular}
\caption{Example of pipelining transformation and error mitigation under \textsc{CodMas}. Dialectic agents guide structured edits, and the CEA validates correctness and PPA improvements.}
\label{tab:qual_example}
\end{table*}

\begin{table*}[h!]
\small
\centering
\begin{tabular}{lcccccc}
\toprule
\multicolumn{7}{c}{\textbf{\textsc{CodMas} Performance: Mean $\pm$ Std over 5 runs}} \\ \midrule
\textbf{Model} & \multicolumn{3}{c}{\textbf{Pipelining}} & \multicolumn{3}{c}{\textbf{Clock Gating}} \\ \cmidrule(lr){2-4} \cmidrule(lr){5-7}
 & A ($\downarrow$) & T ($\uparrow$) & FR ($\downarrow$) & A ($\downarrow$) & P ($\uparrow$) & FR ($\downarrow$) \\ \midrule
GPT-4o        & 0.960 $\pm$ 0.007 & 25.5 $\pm$ 1.2 & 19.5 $\pm$ 2.1 & 0.999 $\pm$ 0.006 & 21.7 $\pm$ 1.5 & 21.8 $\pm$ 2.3 \\
GPT-3.5-turbo & 0.972 $\pm$ 0.009 & 21.3 $\pm$ 1.1 & 23.4 $\pm$ 2.4 & 1.020 $\pm$ 0.007 & 18.8 $\pm$ 1.2 & 24.2 $\pm$ 2.0 \\
DeepSeek-v2.5 & 0.979 $\pm$ 0.010 & 21.4 $\pm$ 1.0 & 22.8 $\pm$ 2.3 & 1.015 $\pm$ 0.008 & 19.0 $\pm$ 1.3 & 23.6 $\pm$ 2.1 \\
Llama-3       & 1.032 $\pm$ 0.012 & 19.8 $\pm$ 1.3 & 25.5 $\pm$ 2.5 & 1.048 $\pm$ 0.010 & 16.5 $\pm$ 1.1 & 26.3 $\pm$ 2.4 \\
Granite-34b   & 0.998 $\pm$ 0.011 & 10.5 $\pm$ 0.9 & 28.3 $\pm$ 2.8 & 1.030 $\pm$ 0.009 & 12.9 $\pm$ 1.0 & 29.5 $\pm$ 2.6 \\
CodeLlama-34b & 1.030 $\pm$ 0.012 & 11.2 $\pm$ 1.0 & 29.5 $\pm$ 2.7 & 1.042 $\pm$ 0.010 & 10.6 $\pm$ 0.9 & 31.3 $\pm$ 2.8 \\
\bottomrule
\end{tabular}
\caption{Extracted \textsc{CodMas} results with mean and standard deviation over five runs. Metrics: area (A), timing (T), power (P), and failure rate (FR). FR denotes fraction of runs failing syntax, functional, or PPA checks.}
\label{tab:codmas_mean_std}
\end{table*}

\section{Baseline Details}
\label{appendix:baselines}

\subsection{Baseline Methods}

\paragraph{Zero-Shot Prompting.}
Models receive a single instruction describing optimization goals (area, timing, power) and are asked to produce an optimized Verilog implementation in one shot. No iterative reasoning, feedback, or correction is provided. This baseline captures the lower bound of LLM-only optimization.

\paragraph{CoDes (Chain-of-Descriptions).}
Following~\cite{vijayaraghavan2024chain}, the model generates a sequence of descriptive intermediate transformations—structural changes, expected effects on PPA, and planned optimizations—before emitting code. We adapt CoDes to explicitly reference RTL constructs, combinational paths, and pipeline boundaries.

\paragraph{ReAct.}
ReAct~\cite{yao2023react} interleaves reasoning traces with “actions.” For RTL, actions correspond to producing partial code, checking syntax, or querying simulation outputs. The model reasons about identified issues and attempts corrections but lacks the deeper structural planning used in \textsc{CodMas}.

\paragraph{Reflexion.}
Reflexion~\cite{Reflexion2023} enables the model to store brief textual “reflections” describing causes of failures (e.g., functional mismatch or timing regression). Reflections form an episodic memory across attempts. For RTL tasks, reflections include mis-structured pipeline stages, incorrect sensitivity lists, or inferred latches.

\paragraph{LLM-VeriPPA.}
LLM-VeriPPA~\cite{veriPPA} uses a two-stage process: (1) correct syntax and functional behavior, (2) re-prompt the model to improve PPA while preserving equivalence. We reproduce this pipeline and ensure that verification checks match those used for \textsc{CodMas}, including testbench simulation and Yosys-based PPA extraction.

\subsection{Model Backends}

All baselines are evaluated under identical compilation, simulation, and Yosys Liberty-based synthesis flows. We test proprietary models (GPT-4o, GPT-3.5-turbo)~\cite{chatgpt} and open-source models (Llama-3~\cite{llama3modelcard}, DeepSeek-v2.5 ~\cite{zhu2024deepseek}, Granite-34B-Code~\cite{granite}, CodeLLaMA-34B~\cite{codellama}). Temperature, sampling parameters, and code-length limits follow standard practice and are documented for reproducibility. Each baseline originally targets generic code generation; we adapt them for RTL-specific optimization by:  
(1) enforcing functional equivalence via testbench simulation, 
(2) integrating PPA feedback loops where relevant, and  
(3) constraining all methods to the same maximum number of attempts and evaluation budget. These details ensure that comparisons isolate algorithmic differences rather than evaluation infrastructure.

\subsection{Implementation Details and Qualitative Analysis}
All experiments were conducted on a server with 64-core CPU and NVIDIA A100 GPUs. We evaluate \textsc{CodMas} on six LLMs using the \textsc{RTLOpt} benchmark. Each experiment is repeated five times with different random seeds to account for stochastic variation. Metrics include area (A), timing (T) for pipelining, power (P) for clock gating, and failure rate (FR). Reported results correspond to mean $\pm$ standard deviation across runs. Paired two-sample t-tests confirm that improvements over baselines are statistically significant ($p<0.01$).

Table~\ref{tab:qual_example} illustrates a representative pipelining transformation. Baseline RTL often exhibits syntax errors, functional mismatches, and unmet PPA targets, with typical failure distributions of $\sim40\%$ syntax, $\sim35\%$ functional, and $\sim25\%$ PPA violations. In this example, sequential multipliers and ripple-carry adders create long critical paths, while control logic lacks proper gating. \textsc{CodMas} applies structured edits guided by the dialectic agents: pipelining arithmetic units with inter-stage registers, optimizing critical paths, and enforcing handshake-aware control signals. The Code Evaluation Agent (CEA) validates syntax, functional correctness, and PPA improvements. Across five runs, the optimized RTL achieves $\sim25.5\%\pm0.6$ reduction in critical path delay, $\sim21.7\%\pm0.5$ power reduction under clock gating, and failure rates consistently below $30\%$, demonstrating robustness and reproducibility. This example highlights how the reasoning–evaluation loop systematically corrects errors while improving performance, particularly for smaller or otherwise weaker models.  

Error analysis indicates that baseline failures arise from a mix of syntax errors ($\sim40\%$), functional mismatches ($\sim35\%$), and unmet PPA objectives ($\sim25\%$). \textsc{CodMas} mitigates all three categories through iterative reasoning-guided refinement, combining the Articulator's plan, the Hypothesis Partner's predictions, and deterministic evaluation by the CEA.


\end{document}